\documentclass[10pt,twocolumn,letterpaper]{article}

\usepackage{cvpr}
\usepackage{times}
\usepackage{epsfig}
\usepackage{graphicx}
\usepackage{amsmath}
\usepackage{amssymb}

\usepackage[breaklinks=true,bookmarks=false]{hyperref}

\cvprfinalcopy 

\def\cvprPaperID{10} 

\ifcvprfinal\fi

\begin{document}

\title{Efficient Plane-Based Optimization of Geometry and Texture for Indoor RGB-D Reconstruction}

\author{Chao Wang\thanks{Author now works at Matterport, Inc.} \ and Xiaohu Guo \\
University of Texas at Dallas\\
800 W Campbell Rd, Richardson, TX 75080, USA\\
{\tt\small \{chao.wang3, xguo\}@utdallas.edu}
}

\maketitle

\begin{abstract}
We propose a novel approach to reconstruct RGB-D indoor scene based on plane primitives. Our approach takes as input a RGB-D sequence and a dense coarse mesh reconstructed from it, and generates a lightweight, low-polygonal mesh with clear face textures and sharp features without losing geometry details from the original scene. Compared to existing methods which only cover large planar regions in the scene, our method builds the entire scene by adaptive planes without losing geometry details and also preserves sharp features in the mesh. Experiments show that our method is more efficient to generate textured mesh from RGB-D data than state-of-the-arts.
\end{abstract}

\section{Introduction}

Online and offline RGB-D reconstruction techniques are developing fast in recent years with the prevalence of consumer depth cameras. State-of-the-art online 3D reconstruction methods can capture indoor and outdoor scenes in the real-world environments efficiently with geometry details \cite{niessner2013real,whelan2015elasticfusion,whelan2016elasticfusion,dai2017bundlefusion,prisacariu2017infinitam}. However, resulting 3D models of these methods are usually too dense with unsatisfying textures due to many reasons including noisy depth data, incorrect camera poses and oversmoothing in data fusion. These models can not be used directly in most applications without further refinement or post-processing.

In order to lower the density and improve the structure quality of indoor models, one typical strategy is to introduce plane primitives into front-end (such as camera tracking in SLAM or online reconstruction in \cite{dou2012exploring, hsiao2017keyframe, halber2017fine}) or back-end (such as RGB-D mesh and texture refinement in \cite{dzitsiuk2017noising, huang20173dlite, wang2018plane}) of reconstruction pipeline, as typical indoor scenes are primarily composed of planar regions, especially buildings and houses with structure following Manhattan-world assumption. However, almost all methods take into account only large planar regions such as walls, ceilings, floors and large table surfaces, and simply ignore and remove other objects with free form surfaces no matter if they contain planar regions or not, such as various indoor furniture and objects on or adhering to large planes. Models with only large planes are too simplified and lack fidelity that they are not applicable to many situations acquiring geometry details. Moreover, geometry details are usually noisy because of noisy RGB-D raw data, and it is difficult and also time-consuming to extract plane primitives or other types of geometry priors from the scene while still preserving the original shape. Besides this, existing back-end methods are usually time-consuming and take hours to process a single scan.

In this paper, we present a novel approach to efficiently reconstruct RGB-D indoor scene using planes and generate a lightweight and complete 3D textured model without losing geometry details. Our method takes as input a RGB-D sequence of indoor scene and a dense coarse mesh reconstructed by some online reconstruction method on this sequence. We firstly partition the entire dense mesh into different planar clusters (Section \ref{sec:partition}), and then simplify the dense mesh into a lightweight mesh without losing geometry details (Section \ref{sec:simp}). Next we create texture patch for each plane and sample points on the plane, and run a global optimization process to maximize the photometric consistency of sampled points across frames by optimizing camera poses, plane parameters and texture colors (Section \ref{sec:opt:plane}). Finally, we optimize the mesh geometry by maximizing consistency between geometry and plane primitives, which further preserves sharp features of original scene such as edges and corners of plane intersections (Section \ref{sec:opt:geo}).  

Our method is highly based on Wang and Guo's method in \cite{wang2018plane}. Compared to their method, the contribution of our method is to introduce line constraint into both pose-plane-texture and geometry optimization, and this can preserve sharp feature better. Meanwhile, our method is also more efficient than \cite{wang2018plane} by introducing parallel computation into the optimization. Experiments show that our method exceeds state-of-the-arts in keeping geometry details and sharp features in the result lightweight 3D textured models. 
\section{Plane-based reconstruction pipeline}

Our reconstruction pipeline takes a RGB-D sequence as input, and firstly uses some state-of-the-art online reconstruction such as VoxelHashing \cite{niessner2013real} or BundleFusion \cite{dai2017bundlefusion} to reconstruct an initial dense mesh.

\subsection{Mesh planar partition}
\label{sec:partition}

We aim to partition the entire mesh into plane primitives to include all geometry details. In our approach we follow the same idea of \cite{wang2018plane} to refer to a state-of-the-art surface partition algorithm proposed by Cai et al. \cite{cai2017surface}. This method proposes a new principle component analysis (PCA) based energy, whose minimization leads to an optimal piecewise-linear planar approximation of the entire surface with high quality. After an input mesh is partitioned into clusters, each cluster is attached with a plane proxy defined by the centroid and normal as the smallest eigenvector direction from the covariance matrix of the cluster. 

After we get the initial planar partition, we run a further plane merging step to merge adjacent planes together into large ones to reduce noisy bumpy points on planar regions. Here we also follow the same idea in \cite{wang2018plane} to merge adjacent planes only if the angle between their normal directions is small enough, and the average distances between two planes are also small. Besides this, we add an additional rule to merge two neighbor planes if the PCA energy increase after merging is very small compared to the initial energy of one plane. This is for merging one large plane and a small neighbor noisy plane together, such as planes on a bumpy floor.

\subsection{Mesh simplification}
\label{sec:simp}

We simplify the mesh based on clusters to create a lightweight mesh for further optimization. Even though we already have a model composed of planes, we still choose to create a mesh by simplifying the original dense mesh instead of using some mesh generation algorithm (such as Delaunay triangulation) on planes like \cite{chauve2010robust, huang20173dlite, li2011globfit}, since it is difficult and also time-consuming to create correct connectivity from complicated plane interceptions in a noisy model, especially an indoor reconstruction mesh containing various geometry objects with free-form shapes. Here we also follow the similar way in \cite{wang2018plane} that uses QEM to simplify the inner-cluster edges at first and then all border edges of clusters next. Note that simplification in each cluster is independent with each other, so we run parallel computation on all inner-cluster edges to accelerate the simplification in our experiments.

\subsection{Plane, camera pose and texture optimization}
\label{sec:opt:plane}

Before optimization, we firstly generate an initial texture mapping for all the faces of the mesh. For each cluster, considering that vertices inside this cluster on the mesh are already near co-planar, we simply project these 3D vertices onto the corresponding plane to get a 2D patch, and sample grid points inside the patch boundary to get texel points, and then backproject them to get corresponding 3D texel points. Another thing is about the keyframes selected from RGB-D frames. To reduce time complexity and increase texture quality, we follow the similar idea of color map optimization by Zhou and Koltun \cite{zhou2014color} to select only sharp frames in every interval, and quatify the blurriness of each image with the metric by Crete et al. \cite{crete2007blur}.

The input in our optimization process is color images $\{\mathbf{I}_i\}$ and depth images of keyframes, all texels' 3D points $\{\mathbf{p}\}$ sampled on the mesh, initial camera poses $\mathbf{T} = \{\mathbf{T}_i\}$ (global to camera space) and initial plane parameters $\Phi = \{\mathbf{\phi}_j\}$. During the optimization, we maximize the photo consistency of 3D texels' projections on corresponding planes across frames by optimizing camera poses, plane parameters and texture colors by minimizing the objective function
\begin{equation}
E_{tex}(\mathbf{T}, \Phi, \mathbf{C}) = E_{c}(\mathbf{T}, \Phi, \mathbf{C}) + \lambda_p E_{p}(\Phi) + \lambda_t E_{t}(\mathbf{T, \Phi}).
\label{eq:main}
\end{equation}
where $\lambda_p$ and $\lambda_t$ are constants to balance different terms.

\textbf{Photometric consistency term.} The photometric energy is designed to measure the photometric error between color of each texel's projection point on its corresponding plane and its target color across frames:
\begin{equation}
E_{c}(\mathbf{T}, \Phi, \mathbf{C}) = \sum_i \sum_{\mathbf{p} \in \mathbf{P}_i} ||C(\mathbf{p}) - \mathbf{I}_i(\pi(\mathbf{T}_i\mathbf{q}))||^2,
\label{eq:color}
\end{equation}
where $C(\mathbf{p})$ is the target color for $\mathbf{p}$, and $\mathbf{P}_i$ is set of all visible 3D texels in frame $i$, and $\pi$ is the perspective projection from 3D position $\mathbf{v}$ to 2D color image, and $\mathbf{q}$ in Eq. (\ref{eq:color}) is the projection point from $\mathbf{p}$ onto its corresponding plane $\phi(\mathbf{p})$ represented by 3D normal $\mathbf{n}_{\mathbf{p}}$ and a scalar $w_{\mathbf{p}}$:
\begin{equation}
\mathbf{q} = \mathbf{p} - (\mathbf{p}^\top \mathbf{n}_{\mathbf{p}} + w_{\mathbf{p}})\mathbf{n}_{\mathbf{p}},
\label{eq:proj}
\end{equation}

\textbf{Plane constraint term.} Plane constraint term is to minimize the sum of distances from 3D texel points to their corresponding planes:
\begin{equation}
E_{p}(\Phi) = \sum_{\mathbf{p}}||\mathbf{p}^\top \mathbf{n}_{\mathbf{p}} + w_{\mathbf{p}}||^2
\end{equation}

\textbf{Line constraint term.} We want to maximize the consistence between 2D lines and corresponding 3D lines which are borders of adjacent planes:
\begin{equation}
E_{t}(\mathbf{T, \Phi}) = \sum_{\mathbf{t} \in \Omega}||(\mathbf{T_{t}^{-1}}\pi^{-1}(\mathbf{t}))^\top \mathbf{n}_{\mathbf{t}} + w_{\mathbf{t}}||^2
\end{equation}
where $\Omega$ is the 2D pixel set with all valid candidate line segments, $\pi^{-1}$ is inverse perspective function of $\pi$ from 2D to 3D. $\Omega$ is obtained by projecting valid 3D line composed of the border vertices shared by clusters onto corresponding visible color image, and then finding its nearest 2D line segment which are computed by line segment detector (LSD) \cite{von2010lsd} within a valid distance range. In experiments we only use 2D line segments with long enough length, and only use 3D lines shared by two planes between which the normal direction is large enough.

To minimize the objective function in Eq. (\ref{eq:main}), we alternate between optimizing different variables with some others fixed and use standard Gauss-Newton method. The optimization of each plane is independent with others so we can solve them in parallel.  Compared to \cite{wang2018plane}, our method removes the image correction term in Eq. (\ref{eq:main}), since we found that it influences very little to the result but highly increases the time complexity with more than 700 additional image correction parameters to optimize per frame. Meanwhile, we add the line constraint term $E_{t}$ to better preserve sharp features.

\begin{figure}[h!]
\begin{center}
\includegraphics[width=0.48\textwidth]{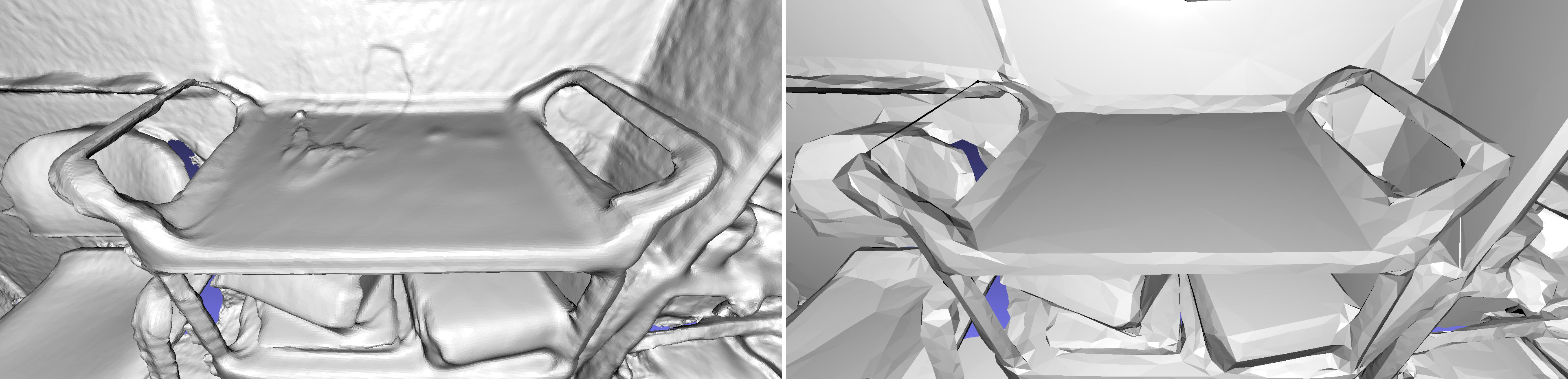}
\end{center}
\caption{Fused dense mesh from BundleFusion (left) and our simplified mesh after optimization with sharp edges (right).}
\label{fig:sharp}
\end{figure}

\begin{figure}[h!]
\begin{center}
\includegraphics[width=0.48\textwidth]{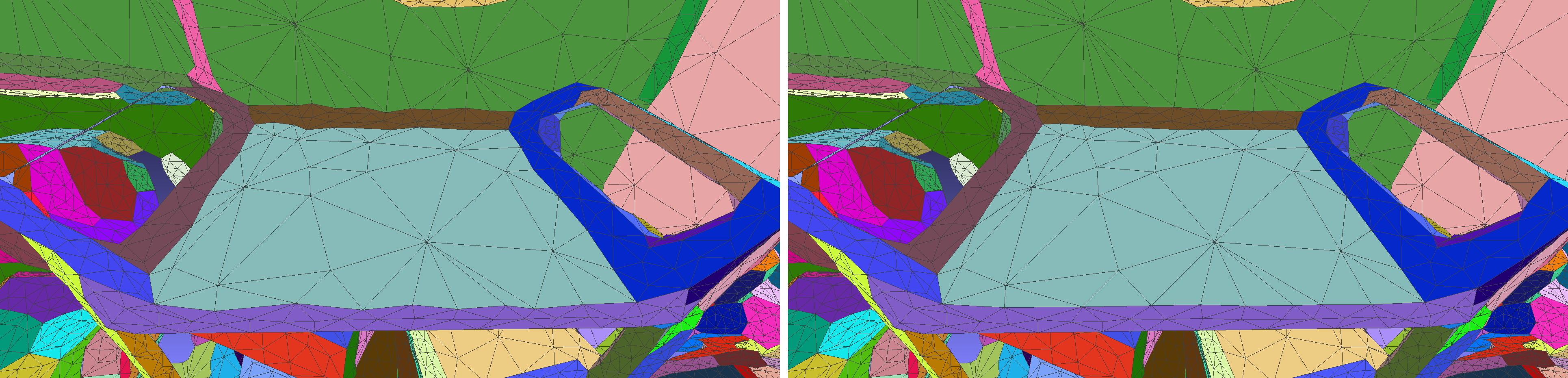}
\end{center}
\caption{Mesh result without (left) or with (right) line constraint terms $E_{t}$ and $E_{l}$. Each color is one cluster.}
\label{fig:mesh}
\end{figure}

\subsection{Geometry optimization}
\label{sec:opt:geo}

The final step is to optimize the mesh geometry to fit the planes as close as possible to reduce noise from mesh surface and sharpen geometry features, since fused meshes reconstructed from RGB-D data always contain noise or oversmoothed surfaces, such as bumpy surfaces on planar regions and smoothed borders which suppose to be sharp features. 

In order to optimize the consistency between geometry and planes, we maximize the consistency between mesh vertices in each cluster and their corresponding planes. Each 2D texel is located inside a triangular face's projection. We utilize the initial barycentric relationship between each texel and its corresponding face, and try to preserve this relationship between texel points' projections on planes and the optimized vertices in each face:
\begin{equation}
E_{vert}(\mathbf{V}) = E_g(\mathbf{V}) + \lambda_l E_{l}(\mathbf{V}) + \lambda_r E_{r}(\mathbf{V}),
\label{eq:geo}
\end{equation}
where $E_g$ is the geometry consistency term
\begin{equation}
E_g(\mathbf{V}) = \sum_{\mathbf{p}}||\mathbf{q} - \sum_{i=0}^2 b_{\mathbf{p},i}\mathbf{v}_{f_{\mathbf{p}}, i}||^2,
\end{equation}
where $\mathbf{q}$ is the projection from 3D texel point $\mathbf{p}$ onto its corresponding plane as described in Eq. (\ref{eq:proj}), $f_{\mathbf{p}}$ is index of the face $\mathbf{p}$ corresponds to, $\mathbf{v}_{f_{\mathbf{p}}, i}$ is the $i$th vertex of face $f_{\mathbf{p}}$, and $b_{\mathbf{p},i}$ is $\mathbf{p}$'s initial barycentric coordinate corresponding to the $i$th vertex in face $f_{\mathbf{p}}$, and $\lambda_l$ and $\lambda_r$ are constants to balance different terms.

\begin{figure*}[h!]
\begin{center}
\includegraphics[width=0.9\textwidth]{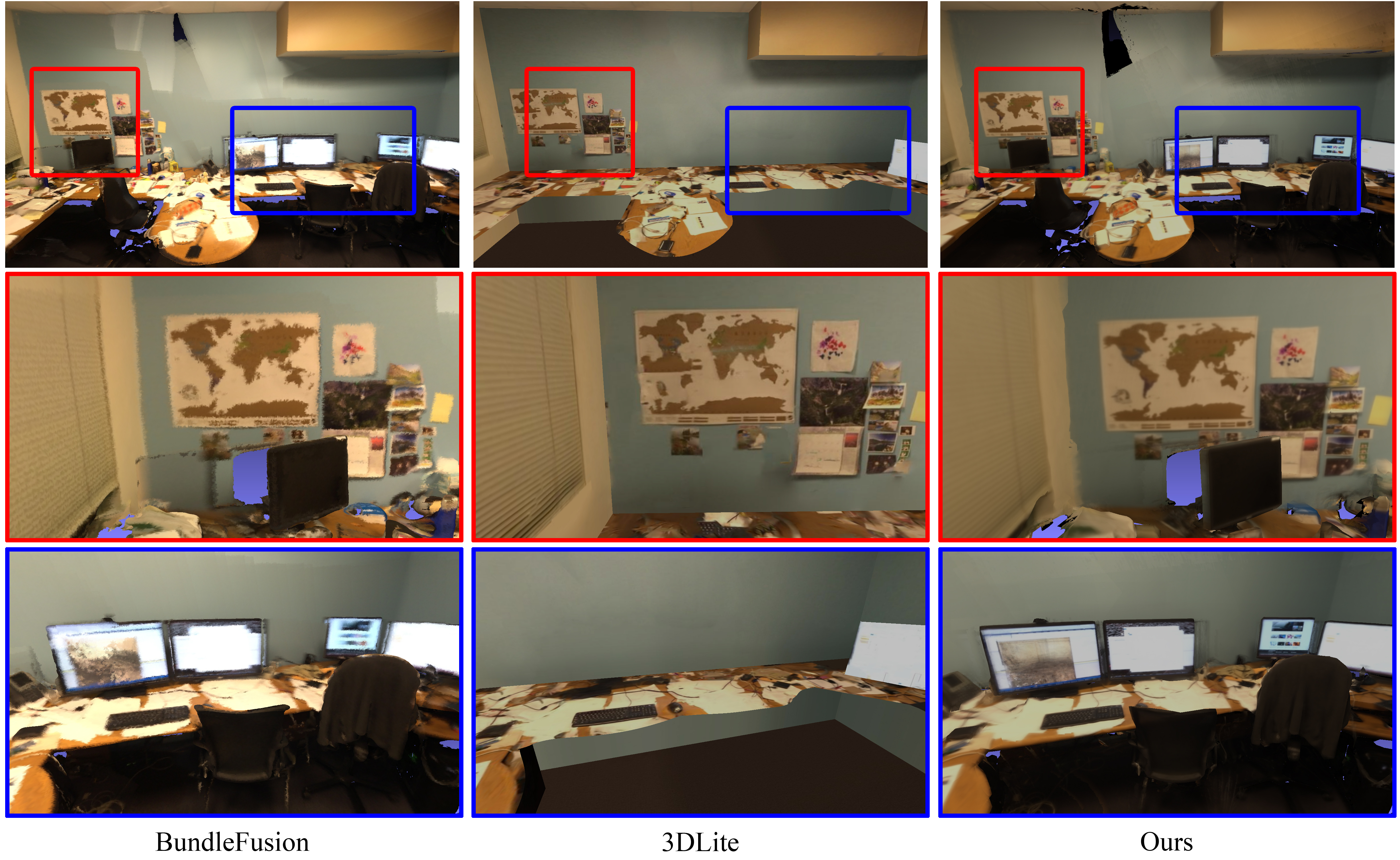}
\end{center}
\caption{Comparison of textured meshes on scan `office3' by BundleFusion \cite{dai2017bundlefusion}, 3DLite \cite{huang20173dlite} and our method. }
\label{fig:comp}
\end{figure*}

Compared to \cite{wang2018plane}, we add a new term $E_{l}$ which is similar to term $E_{t}$ from Eq. (\ref{eq:main}) that it is to ensure all border vertices shared by adjacent planes/clusters are as close to their corresponding planes as possible:
\begin{equation}
E_{l}(\mathbf{V}) = \sum_{p \in \Psi}||\mathbf{p}^\top \mathbf{n}_{\mathbf{p}} + w_{\mathbf{p}}||^2
\end{equation}
where $\Psi$ is the border vertex set. 

The last term $E_r$ in Eq. (\ref{eq:geo}) is a regularization term to minimize the difference between each vertex and the mass center of all its neighbors:
\begin{equation}
E_r(\mathbf{V}) = ||\mathbf{LX}||_F^2.
\end{equation}
Here $\mathbf{X} = [\mathbf{v}_1, \mathbf{v}_2, \cdots, \mathbf{v}_n]^\top$ is matrix of target vertices we want to compute, with $n$ the number of vertices on the mesh. $\mathbf{L}$ is $n\times n$ matrix denoting the discrete graph Laplacian matrix based on the connectivity of the mesh. That is, we want to minimize the difference between each optimized vertex and the average of its neighbor vertices. This term is added to ensure that problem in Eq. (\ref{eq:geo}) has valid solutions. 

The problem in Eq. (\ref{eq:geo}) is actually a sparse linear system and can be solved by Cholesky decomposition efficiently. Figure \ref{fig:sharp} shows comparison between original dense mesh by BundleFusion \cite{dai2017bundlefusion} and our mesh on a scan `office0' from BundleFusion dataset. Our method can preserve the sharp features in the final lightweight mesh very well. Compared to the method in \cite{wang2018plane}, we add another line constraint term $E_l$ to better preserve the line features. Figure \ref{fig:mesh} shows result mesh comparison with or without line constraints on the same scene as Figure \ref{fig:sharp}. 

\section{Results}

We tested our method on the same 10 scans in \cite{wang2018plane} from three popular RGB-D dataset: 6 models from BundleFusion \cite{dai2017bundlefusion} (the first 6 rows in Table \ref{tables}), 2 from ICL-NUIM \cite{handa:etal:ICRA2014} (the following 2 rows ) and 2 from TUM RGB-D dataset \cite{sturm12iros} (the last 2 rows). Table \ref{tables} shows quantitative data of each scan and our result models. Note that the number of faces or vertices of each result model is only 1\%-3\% of that of original dense model. Figure \ref{fig:comp} shows textured mesh between our method and two state-of-the-art systems: BundleFusion \cite{dai2017bundlefusion} and 3DLite \cite{huang20173dlite}, while more strictly speaking, the dense models by BundleFusion are the input of both 3DLite and our method.

We implemented our method in C++\footnote{Result models and part of source code are already open in: \url{https://github.com/chaowang15/plane-opt-rgbd}.} and tested on a desktop with Intel Core i7 2.5GHz CPU and 16 GB memory. The running time on each scan is in Table \ref{tables}. Our average running time is only about 5-10 minutes compared to the average time of several hours in 3DLite \cite{huang20173dlite}, and approximately 30 minutes in Wang and Guo's method \cite{wang2018plane} on the same dataset. We use OpenMP on simplifying different clusters and GPU in computing the Jacobian matrix in plane, texture and pose optimization for acceleration.

\begin{table}[h!]
\begin{center}
\caption{Quantitative data of RGB-D scans and our results. Here $|V|$ is number of vertices, $|F|$ is the number of faces, $|K|$ is the number of keyframes, $t$ is the total running time of our entire pipeline in seconds excluding data I/O.}
\begin{tabular}{|c|c|c|c|c|c|c|}
\hline
Scan & \multicolumn{3}{c|}{Input} & \multicolumn{3}{c|}{Result}\\
\cline{2-7}
 & $|V|$ & $|F|$ & $|K|$ & $|V|$ & $|F|$ & $t$(s) \\
\hline
copyroom & 3.70M & 7.28M & 895 & 55.2K & 104K & 450 \\
\hline
apt0 & 7.83M & 15.4M & 860 & 84.6K & 160K & 521 \\
\hline
office0 & 5.71M & 11.3M & 616 & 68.5K & 130K  & 362 \\
\hline
office1 & 6.03M & 11.9M & 573 & 69.1K & 129K  & 329 \\
\hline
office2 & 5.63M & 11.0M & 700 & 73.6K & 135K & 467 \\
\hline
office3  & 6.36M & 12.6M & 763 & 56.7K & 108K & 485 \\
\hline
of kt2  & 1.20M & 2.36M & 176 & 14.9K & 27.4K & 244 \\
\hline
lr kt2n & 
1.14M & 2.25M & 176 & 22.1K & 41.9K  & 251 \\
\hline
fr2/desk & 1.37M & 2.69M & 372 & 37.6K & 73.4K  & 187 \\
\hline
fr3/loh & 2.42M & 4.75M & 243 & 43.0K & 83.7K  & 126 \\
\hline
\end{tabular}\label{tables}
\end{center}
\end{table}

\textbf{Limitations}. Our method has some similar limitations as \cite{wang2018plane}. Firstly, our face textures are not as sharp as 3DLite's since the latter introduces many techniques to optimize texture, such as texture sharpening and color correction across frames. However, these steps are very time-consuming and possibly takes hours in \cite{huang20173dlite}, and we plan to find a faster way to further optimize textures with similar results as 3DLite. Moreover, our method still cannot fill holes and gaps that always appears in the RGB-D scans, while 3DLite can generate a complete geometry from extracted large planes by extrapolating existing planes and filling holes.

{\small
\bibliographystyle{ieee_fullname}
\bibliography{egbib}
}

\end{document}